%% file: 0-main.tex
\documentclass[sigconf]{acmart}

\usepackage{multirow}
\usepackage[ruled,vlined,linesnumbered]{algorithm2e}
\SetEndCharOfAlgoLine{}
\usepackage{enumitem}
\usepackage{etoolbox}

\usepackage{subfigure}
\usepackage{colortbl}

\usepackage{tabularx}
\usepackage{rotating}

\usepackage{amsthm,amsmath}
\usepackage{mathrsfs}

\usepackage{enumerate}

\usepackage{booktabs}
\usepackage{array}
\usepackage{makecell}
\usepackage{colortbl}
\usepackage{hhline}
\usepackage{graphicx} 
\usepackage{bm}

\AtBeginDocument{%
  \providecommand\BibTeX{{%
    \normalfont B\kern-0.5em{\scshape i\kern-0.25em b}\kern-0.8em\TeX}}}

\copyrightyear{2025}
\acmYear{2025}
\setcopyright{rightsretained}

\acmConference[CIKM '25] {Proceedings of the 34th ACM International Conference on Information and Knowledge Management}{ November 10--14, 2025}{Seoul, Republic of Korea.}
\acmBooktitle{Proceedings of the 34th ACM International Conference on Information and Knowledge Management (CIKM '25), November 10--14, 2025, Seoul, Republic of Korea}
\acmISBN{979-8-4007-2040-6/2025/11}
\acmDOI{10.1145/XXXXXX.XXXXXX}
\makeatletter
\patchcmd{\maketitle}{\@copyrightpermission}{
   \begin{minipage}{0.3\columnwidth}
     \href{https://creativecommons.org/licenses/by/4.0/}{\includegraphics[width=0.90\textwidth]{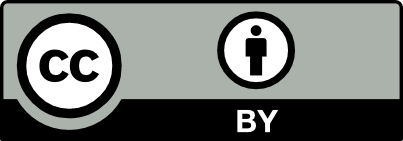}}
   \end{minipage}\hfill
   \begin{minipage}{0.7\columnwidth}
     \href{https://creativecommons.org/licenses/by/4.0/}{This work is licensed under a Creative Commons Attribution International 4.0 License.}
   \end{minipage}
   \vspace{5pt}
}{}{}
\makeatother

\begin{document}

\author{Kun Peng}
\affiliation{%
  \institution{Institute of Information Engineering, Chinese Academy of Sciences \& School of Cyber Security,
 University of Chinese Academy of Sciences,}
  \city{Beijing}
  \country{China}}
\email{pengkun@iie.ac.cn}

\author{Cong Cao}
\affiliation{%
  \institution{Institute of Information Engineering,
 Chinese Academy of Sciences,}
  \city{Beijing}
  \country{China}}
\email{caocong@iie.ac.cn}
\authornote{Corresponding author}

\author{Hao Peng}
\affiliation{%
  \institution{School of Cyber Science and Technology, Beihang University,}
  \city{Beijing}
  \country{China}
}
\email{penghao@buaa.edu.cn}

\author{Zhifeng Hao}
\affiliation{%
  \institution{College of Science, University of Shantou,}
  \city{Shantou}
  \country{China}
}
\email{haozhifeng@stu.edu.cn}

\author{Lei Jiang}
\affiliation{%
  \institution{Institute of Information Engineering,
 Chinese Academy of Sciences,}
  \city{Beijing}
  \country{China}}
\email{jianglei@iie.ac.cn}

\author{Kongjing Gu}
\affiliation{%
  \institution{National University of Defense Technology,}
  \city{Changsha}
  \country{China}}
\email{gu_kongjing@outlook.com}

\author{Yanbing Liu}
\affiliation{%
  \institution{Institute of Information Engineering, Chinese Academy of Sciences \& School of Cyber Security,
 University of Chinese Academy of Sciences,}
  \city{Beijing}
  \country{China}}
\email{liuyanbing@iie.ac.cn}

\author{Philip S. Yu}
\affiliation{%
  \institution{Department of Computer Science, University of Illinois at Chicago,}
  \city{Chicago, Illinois,}
  \country{USA}}
\email{psyu@uic.edu}

\renewcommand{\shortauthors}{Kun Peng et al.}

\theoremstyle{definition}
\newtheorem{define}{Definition}[]

\title{Dialogues Aspect-based Sentiment Quadruple Extraction via Structural Entropy Minimization Partitioning}

\begin{abstract}
Dialogues Aspect-based Sentiment Quadruple Extraction (DiaASQ) aims to extract all target-aspect-opinion-sentiment quadruples from a given multi-round, multi-participant dialogue. Existing methods typically learn word relations across entire dialogues, assuming a uniform distribution of sentiment elements. However, we find that dialogues often contain multiple semantically independent sub-dialogues without clear dependencies between them. Therefore, learning word relationships across the entire dialogue inevitably introduces additional noise into the extraction process. To address this, our method focuses on partitioning dialogues into semantically independent sub-dialogues. Achieving completeness while minimizing these sub-dialogues presents a significant challenge. Simply partitioning based on reply relationships is ineffective. Instead, we propose utilizing a structural entropy minimization algorithm to partition the dialogues. This approach aims to preserve relevant utterances while distinguishing irrelevant ones as much as possible. Furthermore, we introduce a two-step framework for quadruple extraction: first extracting individual sentiment elements at the utterance level, then matching quadruples at the sub-dialogue level. Extensive experiments demonstrate that our approach achieves state-of-the-art performance in DiaASQ with much lower computational costs.
\end{abstract}




\maketitle
\input{1-intro}
\input{2-relatedwork}
\input{4-model}
\input{5-experiments}
\input{6-conclusions}



\bibliographystyle{ACM-Reference-Format}
\bibliography{sample-base}


\end{document}

%% file: 1-intro.tex
\section{Introduction}\label{sec:introduction}
\begin{figure}[t]
    \hspace{-0.1cm}
    \centering
    \includegraphics[width=0.49\textwidth]{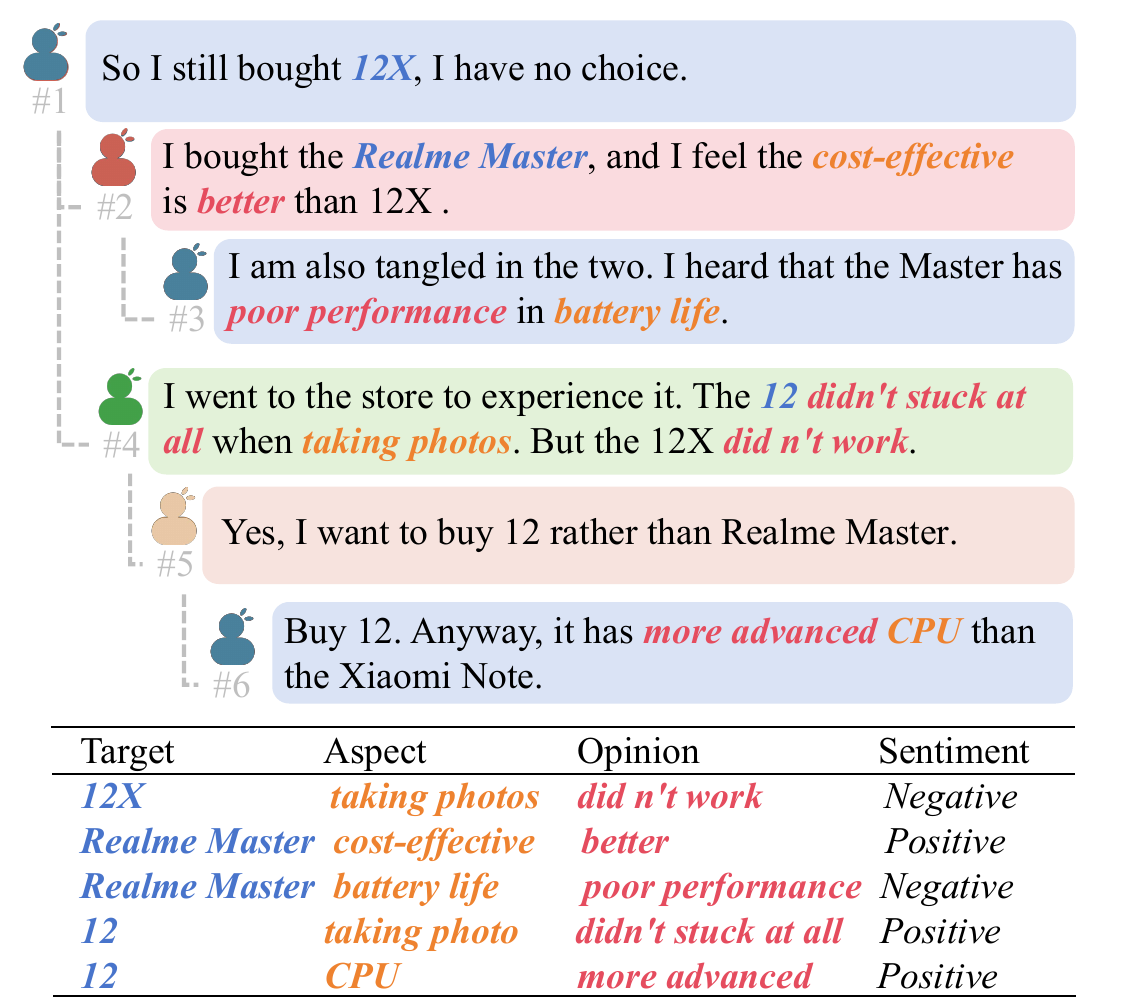}
    \caption{An example of the DiaASQ task. This dialogue has six utterances with four speakers.}
    \label{fig:0}
\end{figure}

Aspect-based Sentiment Analysis (ABSA) is a focused research direction aimed at extracting specific aspect terms and their associated sentiment elements (e.g., opinion words, sentiment polarities, aspect categories) from an individual sentence \cite{9996141}. 
Recent studies such as Aspect-Opinion Pair Extraction (AOPE) \cite{chen-etal-2020-synchronous, ijcai2021p545}, Aspect Sentiment Triplet Extraction (ASTE) \cite{DBLP:conf/aaai/PengXBHLS20, xu-etal-2021-learning}, and Aspect Sentiment Quad Prediction (ASQP) \cite{zhang-etal-2021-aspect-sentiment, zhou-etal-2023-unified-one} have explored various configurations of extracted sentiment elements.
Despite significant progress, single-sentence ABSA techniques fall short in meeting the demands of real-world multi-turn, multi-participant dialogue sentiment analysis tasks.
To propel this research direction forward, \citet{li-etal-2023-diaasq} proposed the Dialogue Aspect-based Sentiment Quadruple Extraction (DiaASQ) task. 
As illustrated in Figure \ref{fig:0}, DiaASQ aims to extract all \textit{target-aspect-opinion-sentiment} quadruples from a collection of multi-turn dialogue texts.
For a target of commentary in the dialogue, there are multiple aspects of critique, along with opinion words and sentiment polarities towards these aspects.

Compared to traditional ABSA tasks, DiaASQ faces the primary challenge of effectively capturing long-distance sentiment relationships across sentences. 
As depicted in Figure \ref{fig:0}, elements of a sentiment quadruple may be distributed across different utterances, placing more stringent demands on models to adeptly learn these sentiment relationships across sentence boundaries.
Meta-WP \cite{li-etal-2023-diaasq} introduces the use of three attention matrices (i.e. the same thread, the same speaker, and the reply relation) to learn relationships between different words at the scale of the entire dialogue. 
Furthermore, H2DT \cite{Li2024HarnessingHD} constructs a reply-speaker heterogeneous graph and utilizes GCN to thoroughly learn word relationships based on this graph.
However, due to the multi-turn, multi-participant nature of dialogues, these methods learn word relationships at the scale of the entire dialogue, inevitably introducing noise into the capture of local relationships.
Taking the targets \textit{Realme Master} and \textit{12} from Figure \ref{fig:0} as an example, their related sentiment elements belong to different dialogue threads, which means there is no clear semantic dependency between them. 
Therefore, learning word dependencies across the entire dialogue is not necessary.
Furthermore, since different dialogue threads all belong to the same topic and share a common root utterance, there is a semantic similarity that introduces noise into the learning of local word relationships. 
For instance, the occurrence of the target word \textit{Realme Master} in both utterance \#2 and utterance \#5 undoubtedly affects the model's performance.

Based on this motivation, we propose an advanced \textbf{S}tructural \textbf{E}ntropy \textbf{M}inimization-based \textbf{Dia}ASQ model, called \textbf{SEMDia}.
Our core insight is to partition the dialogue texts into minimal, semantically complete sub-dialogue units and extract sentiment quadruples from them. 
Since sentiment quadruples are not uniformly distributed across independent dialogue threads, unwise partitioning methods—such as dividing elements belonging to the same quadruple into different sub-dialogues can lead to decreased effectiveness. 
Therefore, effectively partitioning the dialogue presents the primary challenge.
To achieve this, we begin by utilizing a pre-trained language model to encode each utterance in the dialogue and construct a dialogue graph. 
These utterance embeddings serve as nodes in the graph, where edges with similarity weights connect nodes within the same dialogue thread.
Following the dialogue graph construction, we employ a Dialogue Structural Entropy Minimization (DSEM) algorithm to partition the graph into clusters of utterances that are structurally cohesive, without requiring a predefined number of clusters.
Once we obtain these minimal sub-dialogue clusters, we propose a two-step framework for sentiment quadruple extraction. 
Initially, we extract individual sentiment elements at the utterance level. 
Subsequently, within each sub-dialogue cluster, we match these sentiment elements to form sentiment quadruples.
By reducing noise through the previous clustering process, our two-step extraction framework achieves more refined results.
Our contributions can be summarized as follows:

1) We propose using a dialogue structural entropy minimization (DSEM) algorithm to partition dialogues into semantically complete minimal sub-dialogues, thereby reducing additional noise introduced when extracting quadruples at the entire dialogue scale. 

2) The sub-dialogues partitioning based on the DSEM algorithm is unsupervised, requiring no additional parameters or prior knowledge of the number of sub-dialogues.

3) Extensive experiments demonstrate that our method achieves state-of-the-art performance with much lower computational costs compared to previous methods.





%% file: 2-relatedwork.tex
\section{Related Work}\label{sec:relatedwork}
\subsection{Aspect-based Sentiment Analysis}
Compared to traditional coarse-grained sentiment analysis, Aspect-based Sentiment Analysis (ABSA) shifts its focus to extracting sentiment specifically towards entities or their specific aspects.
Depending on the extracted sentiment elements, ABSA encompasses various subtasks, one of the most related subtasks to DiaSQE is Aspect Sentiment Triplet Extraction (ASTE) \cite{DBLP:conf/aaai/PengXBHLS20, xu-etal-2021-learning, Peng2024PromptBT, 10.1145/3627673.3679543}, which focuses on extracting triplets consisting of aspect terms, opinion words, and sentiment polarity.
Early works \cite{zhang-etal-2022-boundary, Liang2022STAGEST} were based on the Grid Tagging Scheme (GTS) \cite{wu-etal-2020-grid} approach, extracting sentiment elements by tagging token pairs. Some other works \cite{xu-etal-2021-learning, chen-etal-2022-span} focus on the display interaction of sentiment elements at the span level. These methods are all dedicated to learning comprehensive token dependencies.
In another line, generation-based methods have also shown promise, \citet{zhai-etal-2022-com} treated the task as a reading comprehension problem, \citet{gou-etal-2023-mvp} designed prompt templates to guide model generation.

Although ABSA has been well-studied at the sentence level, it struggles to achieve deeper success in dialogue scenarios, especially with the vast amount of product-related discussions on social platforms.

\subsection{Dialogues Aspect-based Sentiment Quadruple Extraction}
Previous works \cite{li-etal-2020-modeling, hu-etal-2021-dialoguecrn} made initial explorations into dialogue-level sentiment analysis, but their efforts remain at a coarse granularity.
Recently, \citet{DBLP:journals/jair/SongXLWSX22} proposed a fine-grained task of extracting dialogue-level sentiment triplets. Furthermore, \citet{li-etal-2023-diaasq} introduced the task of extracting dialogue-level sentiment quadruples, named DiaASQ.
\citet{Li2024HarnessingHD} addressed the DiaASQ task by using GCN to model dialogue structure.
\citet{10447873} introduced a multi-scale aggregation network to fully aggregate contextual information.
\citet{huang-etal-2024-dmin} employed a multi-granularity integrator to enhance token-level contextual understanding.
These advanced works all focused on aggregating information at the entire dialogue level. However, we believe this may also introduce more noise, and a reasonable partitioning of the dialogue would be beneficial.

%% file: 4-model.tex
\section{Preliminary}\label{sec:preliminary}

\begin{figure*}[t]
    \centering
    \includegraphics[width=0.95\textwidth]{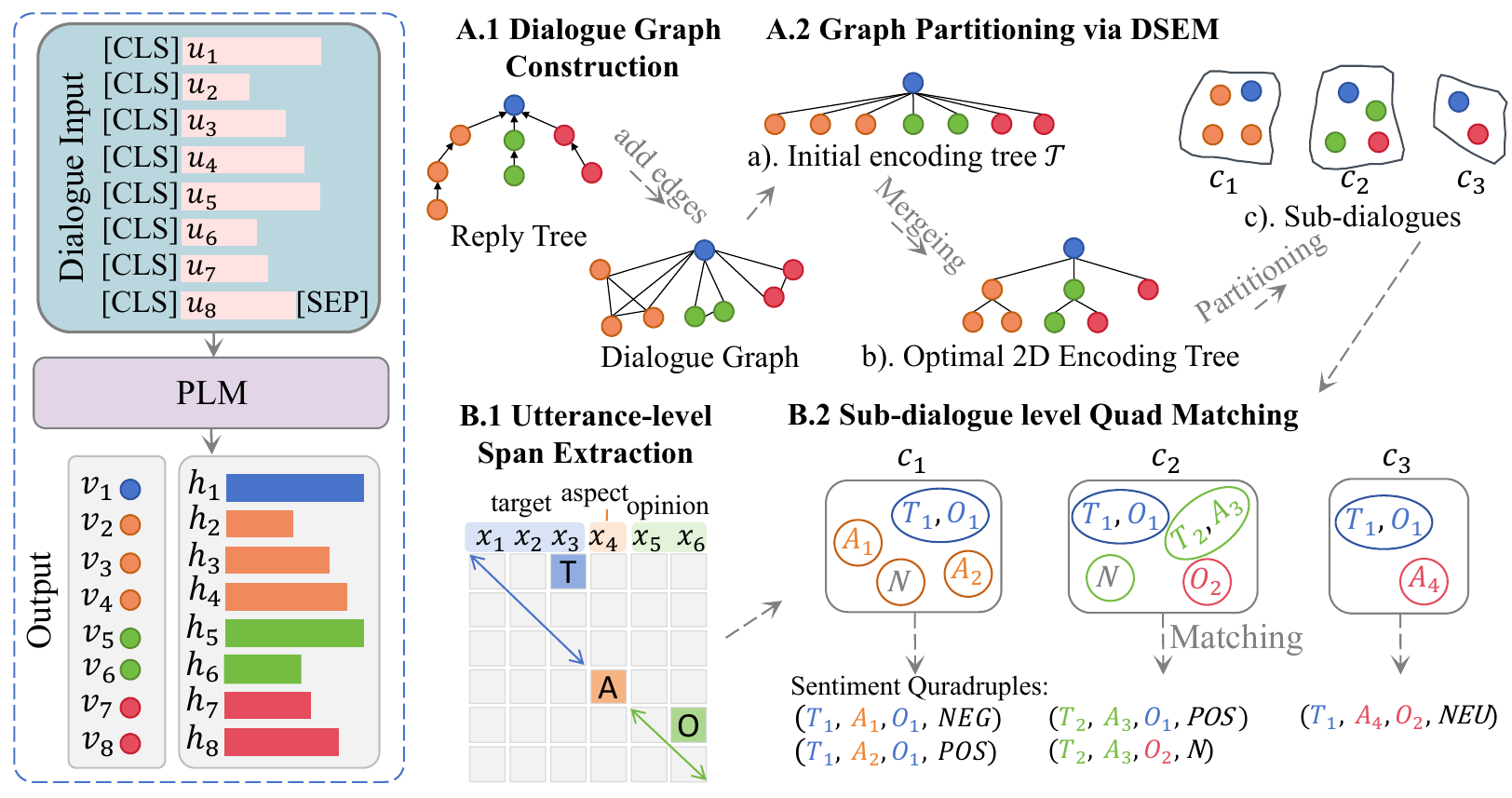}
    \vspace{-0.2cm}
    \caption{The framework of our proposed SEMDia. SEMDia consists of two phases.
In Phase A., we employ a two-step graph construction-partitioning strategy to partition the dialogue into several smaller sub-dialogues.
In Phase B., we first propose an utterance-level span extraction method, and then propose a quadruple matching method at the sub-dialogue level.}
    \label{fig:1}
\end{figure*}

Structural Entropy (SE) theory \cite{DBLP:journals/tit/LiP16, Cao_Peng_Yu_Yu_2024, 10.1145/3660522} was originally introduced to measure the complexity of a graph's structure.
It is defined as the minimum number of bits required to encode a vertex accessible through random walks on the graph.
Specifically, given a homogeneous graph $G = (\mathcal{V}, \mathcal{E})$ and undergoing hierarchical partition, the underlying substantive structure is obtained by minimizing the structural entropy of the corresponding encoding tree.
We present encoding trees and SE as follows.

\noindent \textbf{Definition 1.} \textit{An encoding tree $\mathcal{T}$ of graph $G$ is defined with the following properties:}
\begin{itemize}[label=(\arabic*)]
\item[\textit{1)}] \textit{Each node $\alpha$ in the encoding tree $\mathcal{T}$ has a associated subset $T_\alpha \subseteq \mathcal{V}$.}

\item[\textit{2)}] \textit{For root node $\lambda$ in $\mathcal{T}$, $T_\lambda = \mathcal{V}$. For any leaf node $\gamma$ in $\mathcal{T}$, $T_\gamma = \{v\}$, where $v$ is a single node in $\mathcal{V}$.
}

\item[\textit{3)}] \textit{The $n$ childrens of node $\alpha$ are denote as $\beta_1,...,\beta_n$, the mutually exclusive ${T_\beta}_1, ...,{T_\beta}_n$ is a partition of $T_\alpha$. }

\item[\textit{4)}] \textit{The height of node $\alpha$ is denoted as $h(\alpha)$. $h(\gamma)=0$ and $h(\alpha^-)=h(\alpha)+1$, where $\alpha^-$ is the parent of $\alpha$.
}

\item[\textit{5)}] \textit{The height of $\mathcal{T}$ is denoted as $h(\mathcal{T})=\mathop{\max}_{\alpha \in \mathcal{T}} h(\alpha)$.
}
\end{itemize}

Intuitively, the encoding tree reflects the hierarchical division of the graph. 
A $k$-dimensional encoding tree means that $h(\mathcal{T})=k$.

\noindent \textbf{Definition 2.} \textit{The SE of graph $G$ determined by the encoding tree $\mathcal{T}$ is defined as:}
\begin{equation}
\mathcal{H}^\mathcal{T}(G)=-\sum_{\alpha \in \mathcal{T}, \alpha \neq \lambda} \frac{g_\alpha}{vol(\lambda)} \log \frac{vol(\alpha)}{vol(\alpha^-)},
\end{equation}
\textit{where $g_\alpha$ is sum of edge weights from $\mathcal{T}_\alpha$ to all external vertices. 
$vol(\lambda)$, $vol(\alpha)$ and $vol(\alpha^-)$ represent the degrees of vertices $\lambda$, $\alpha$ and $\alpha^-$, respectively. }

Depending on the order, a graph $G$ has multiple different $k$-dimensional encoding trees. 
The optimal $k$-dimensional encoding tree is defined as having the minimum SE.
Thus, we define the k-dimensional SE of graph $G$ as: $\mathcal{H}^{(k)}(G)=\mathop{\min}_{\forall \mathcal{T}: h(\mathcal{T})=k}\mathcal{H}^\mathcal{T}(G)$.

\section{Method}\label{sec:method}
The overall architecture of our proposed Structural
Entropy Mini-mization-based DiaASQ (SEMDia\footnote{Codes and data are available at https://github.com/KunPunCN/SEMDia/}) model is shown in Figure \ref{fig:1}. 
In the following part, we will present the task definition in Section \ref{sec:task}, then introduce our SEMDia step by step.

\subsection{Task Definition}
\label{sec:task}
Given a dialogue $D=\{(u_1,s_1), (u_2,s_2), ...,$ $ (u_n,s_n)\}$ and a reply list $L=\{l_1, l_2, ..., l_n\}$, where $(u_i,s_i)$ represent the $i$-th utterance and its speaker, respectively. $n$ is the utterance number in the dialogue, $l_i$ represents that the $i$-th utterance is a reply of the $l_i$-th utterance.
Specifically, the first utterance serves as the root, hence $l_1=-1$. 
The objective of DiaASQ is to extract all sentiment quadruples $ Q = \{(t_i, a_i, o_i, p_i)\}_{i=1}^{|Q|}$ from $D$, where $( t_i, a_i, o_i )$ are spans in the text representing the target span, associated aspect span, and opinion span respectively, and $p_i$ denotes the corresponding sentiment polarity, categorized as \{\textit{positive}, \textit{negative}, \textit{neutral}\}.

\subsection{A.1 Dialogue Graph Construction}
SEMDia first divides the dialogue into minimal sub-dialogues using a graph construction-partitioning strategy.
Step $A.1$ in Figure \ref{fig:1} illustrates the graph construction process.
This step initially encodes utterance representations using a pre-trained language model (PLM), where the input to the PLM is constructed as follows: 
\begin{equation}
Input = [CLS] \oplus u_1 \oplus ...\oplus [CLS] \oplus u_n  \oplus [SEP],
\end{equation}
where $\oplus$ means the concatenation operation.
The output is: 
\begin{equation}
PLM(Input) = v_1 \oplus h_1 \oplus ... \oplus h_n \oplus v_{n+1},
\label{eq3}
\end{equation}
where $\bm{H} = \{h_1, ..., h_n\}$ are the encoded tokens of utterances $u_1, ..., u_n$. We take $\mathcal{V} = \{v_1, ..., v_n\}$ corresponding to each [CLS] as the embedding of each utterance. 

Next, we designate $\mathcal{V}$ as the nodes of dialogue graph $G=\{\mathcal{V}, \mathcal{E}\}$. 
To establish edges $\mathcal{E}$ between nodes, given the reply list $L$, a bidirectional edge is placed between nodes that share a reply relation. 
For any two nodes $v_i$ and $v_j$ not directly connected, if there exists a node $v_k$ such that there is a path $ v_i \rightarrow v_k \rightarrow v_j $, we add a path $ v_i \rightarrow v_j $.
The edge weight $w_{ij}$ is computed as the cosine similarity of the embeddings: 
\begin{equation}
w_{ij} = max(cosine(v_i, v_j), 0).
\end{equation}
Finally, after adding self-loops, we construct an undirected dialogue graph $G$.

\subsection{A.2 Graph Partitioning via DSEM}
By structuring the dialogue into a graph, we can further partition the dialogue using a Structural Entropy Minimization (SEM) algorithm \cite{cao2024hierarchical,10.1145/3660522}.
The pseudo-code of our proposed Dialogue SEM (DSEM) is shown in Algorithm \ref{xxx}.

As illustrated in Figure \ref{fig:1} A.2, we first initialize the encoding tree $\mathcal{T}$, as defined in the Preliminary Section. Each leaf node in the tree $\mathcal{T}$ corresponds to a vertex in the graph $G$. 
The initial one-dimensional encoding tree represents the simplest two-level structure, where the leaf nodes in the graph are directly connected to the root node, and each leaf node is treated as an individual cluster.
The initial division of sub-dialogues is defined as \textit{subs} (line 1). 
If two leaf nodes share the same parent node, it indicates they belong to the same sub-dialogue cluster.

For two subtrees $\mathcal{T}_\alpha$ and $\mathcal{T}_\beta$ in tree $\mathcal{T}$ sharing the same parent node, we can perform a merge operation $Merge(\mathcal{T}_\alpha,\mathcal{T}_\beta)$ to generate a new tree $\mathcal{T}'_{(\alpha,\beta)}$, which can be understood as combining cluster $\beta$ into $\alpha$.
Before and after performing the merge operation, the change in SE of tree $\mathcal{T}$ is calculated as:
\begin{equation}
\Delta(\mathcal{T}; \mathcal{T}_\alpha, \mathcal{T}_\beta) = SE' - SE,
\end{equation}
\begin{equation}
SE’ = \mathcal{H}^\mathcal{\mathcal{T}'_{(\alpha,\beta)}}(G; \alpha) + \sum_{\epsilon^-=\alpha}{\mathcal{H}^\mathcal{\mathcal{T}'_{(\alpha,\beta)}}(G; \epsilon)},
\end{equation}
\begin{equation}
SE = \mathcal{H}^\mathcal{\mathcal{T}}(G; \alpha) + \mathcal{H}^\mathcal{\mathcal{T}}(G; \beta) + \sum_{\epsilon^-=\alpha or \beta}{\mathcal{H}^\mathcal{\mathcal{T}}(G; \epsilon)},
\end{equation}
where $\mathcal{H}^{\mathcal{T}'_{(\alpha,\beta)}}(G; \alpha)$ and $\mathcal{H}^{\mathcal{T}'_{(\alpha,\beta)}}(G; \epsilon)$ are the SE of $\mathcal{T}_\alpha$ and $\mathcal{T}_\epsilon$ (subtrees of $\mathcal{T}_\alpha$) after merge operation, respectively.
$\mathcal{H}^\mathcal{T}(G; \alpha)$, $\mathcal{H}^\mathcal{T}(G; \beta)$ and $\mathcal{H}^\mathcal{T}(G; \epsilon)$ are the SE of $\mathcal{T}_\alpha$, $\mathcal{T}_\beta$ and $\mathcal{T}_\epsilon$ (subtrees of $\mathcal{T}_\alpha$ and $\mathcal{T}_\beta$) before merge operation, respectively.

If $\Delta(\mathcal{T}; \mathcal{T}_\alpha, \mathcal{T}_\beta) > 0$, it indicates a decrease in SE of tree $\mathcal{T}$ after the merge operation, thereby reducing uncertainty.
We iterate all edges in tree $\mathcal{T}$ to obtain the SE change resulting from merging any two subtrees (lines 5-8), selecting the edges corresponding to the top $\sigma$ largest change values, and performing the merge operation for these edges (lines 9-12).
We repeat the above operation until the SE of the tree $\mathcal{T}$ can no longer decrease (lines 13-17).
In the final optimal 2D encoding tree (Figure \ref{fig:1} A.2b), the graph structure reaches a stable state, thereby achieving the optimal partitioning of sub-dialogues.
In particular, because the root utterance contains common topic information, we include it in each sub-dialogue cluster during partitioning (Figure \ref{fig:1} A.2c).
The collection of sub-dialogues is donated as $\mathcal{C}$.

\begin{algorithm}[t]
  \SetAlgoLined
  \SetKwInOut{Input}{Input}
  \SetKwInOut{Output}{Output}
  \Input{The dialogue graph $G=\{\mathcal{V}, \mathcal{E}\}$, the parameters for parallel Merge Operators $\sigma$.}
  \Output{the set of partitioned sub-dialogues \textit{subs}.}
  the initial division \textit{subs} $\leftarrow$ $G$, \textit{current\_num} $\leftarrow$ number of $\mathcal{V}$, \textit{edge} $\leftarrow \mathcal{E}$\;
  \textit{merge\_all} $\leftarrow$ False\;
  
  \While{not \textit{merge\_all}}{
    \textit{max\_operate} $\leftarrow$ (\textit{current\_num}-1) * $\sigma$\;
    
    \For {$(\alpha,\beta)$ in \textit{edge}}{
        \If{$\Delta(\mathcal{T}; \mathcal{T}_\alpha, \mathcal{T}_\beta) > 0$}{
            $sets \stackrel{\text{add}}{\leftarrow} (\Delta(\mathcal{T}; \mathcal{T}_\alpha, \mathcal{T}_\beta), \alpha,\beta)$
        } 
    }
    \If{$len(sets)>\textit{max\_operate}$}{
        sort $sets$ by $\Delta(\mathcal{T}; \mathcal{T}_\alpha, \mathcal{T}_\beta)$\;
        $\textit{op\_edge} \stackrel{\text{add}}{\leftarrow} (\alpha,\beta)$ with top \textit{max\_operate} $\Delta(\mathcal{T}; \mathcal{T}_\alpha, \mathcal{T}_\beta)$\;
        }
    \If{$len(sets)=0$}{
        \textit{merge\_all}=True\;
        }
    \textit{current\_num} $\leftarrow \textit{current\_num}-len(\textit{op\_edge})$\;
    update \textit{edge} and \textit{subs}.
  }
 \caption{Dialogue SE Minimization (DSEM)}
 \label{xxx}
\end{algorithm}

\subsection{B.1 Utterance-level Span Extraction}
For extracting entity spans, it is crucial to determine the correct span boundaries. 
To achieve this, we transform the original one-dimensional utterance embedding into a two-dimensional relation table, where the relationships between any two boundary positions in the utterance are represented in cells of the table.
As shown in Figure \ref{fig:1} B.1, inspired by \cite{Li2024HarnessingHD}, we annotate entities in a two-dimensional table, where the vertical axis denotes the span start, and the horizontal axis denotes the span end.
Therefore, for any entity span in utterance within the range $[a, b]$ (where $a \leq b$), it can be uniquely tagged at coordinates $(a, b)$ in the table.
Hence, there are four distinct labels in total: \{Target ($\textit{T}$), Aspect ($\textit{A}$), Opinion ($\textit{O}$), Invalid ($\textit{Inv}$)\}.

\subsubsection{Entity Relation Table Construction}
Specifically, from Equation \ref{eq3}, we can obtain the representation $h \in \mathbb{R}^{m \times d}$ from $\bm{H}$ for any utterance, where $m$ is the length of the utterance and $d$ the dimension of the PLM's hidden layer.
Then, two dense MLPs are employed to learn the start and end boundaries of the utterance: 
\begin{equation}
h^{s} = MLP_s(h), h^{e} = MLP_e(h), 
\label{eq7}
\end{equation}
where $h^{s}, h^{e} \in \mathbb{R}^{m \times d'}$ are boundary representations.
After this, we construct the boundary relation table $R \in \mathbb{R}^{m \times m \times (3d'+\eta)}$ using the following formula:
\begin{equation}
r_{i,j} = h^{s}_i \oplus h^{e}_j \oplus Pool(h^{s}_{[i:j]}) \oplus Biaf(h^{s}_i, h^{e}_j),
\end{equation}
\begin{equation}
Pool(h_{[i,j]}) = Maxpooling(h_i,...,h_j),
\end{equation}
\begin{equation}
Biaf(h_i, h_j) = {h_i}^T \mathbf{W_1} h_j,
\label{eq10}
\end{equation}
where $r_{i,j} \in \mathbb{R}^{3d'+\eta}$ is the $(i,j)$-th representation of entity relation table $R$. 
$Pool(\cdot)$ is a max-pooling operation, which aims to obtain the entire information of the span range $[i,j]$ in the utterance.
$Biaf(\cdot)$ is a Biaffine Attention \cite{dozat2017deep} operation and $\mathbf{W_1} \in \mathbb{R}^{d' \times \eta \times d'}$ is a parameter matrix.
To further enhance the boundary relations, we input $R$ into a dense layer followed by a single CNN layer, obtaining the final entity table representation $\widetilde{R} \in \mathbb{R}^{m \times m \times d'}$: 
\begin{equation}
\widetilde{r}_{i,j} = Conv_{(3\times 3)}(MLP_d(r_{i,j})),
\label{eq10}
\end{equation}
where $MLP_d: \{\mathbb{R}^{3d'+\eta} \rightarrow \mathbb{R}^{d'} \}$ is the dense layer.
$Conv_{(3\times 3)}$ is a single CNN layer with kernel size $(3 \times 3)$ and stride size 1.

\subsubsection{Entity Extraction}
Finally, we employ a label-wise linear layer $MLP_{ent}: \{\mathbb{R}^{d'} \rightarrow \mathbb{R}^{4} \}$ to predict the label ${p}^{ent}_{i,j} \in \{ \textit{T}, \textit{A}, \textit{O}, \textit{Inv}\}$ for each position in $\widetilde{R}$: 
\begin{equation}
{p}^{ent}_{i,j} = MLP_{ent}({\widetilde{r}_{i,j}}).
\end{equation}
If ${p}^{ent}_{i,j} $ belongs to $ \{ \textit{T}, \textit{A}, \textit{O}\}$, then the span within the range $[i,j]$ can be extracted as the corresponding entity; otherwise, if classified as \textit{Inv}, it is filtered out.
The loss of the span extraction module is formulated as follows:
\begin{equation}
l^{ent} = -\frac{1}{m^2} \sum_{i=1}^m \sum_{j=1}^m \log ({p}^{ent}_{i,j} I({y}^{ent}_{i,j})),
\end{equation}
where $l^{ent}$ is the loss of one utterance with length $m$, ${y}^{ent}_{i,j} \in \{ \textit{T}, \textit{A}, \textit{O}, \textit{Inv}\}$ is the label of position $(i,j)$. 
$I(\cdot)$ is an indicator function.
The sum of the losses $l^{ent}$ for all utterances is calculated as: 
\begin{equation}
\mathcal{L}_{ent} = \sum_{i=1}^n l^{ent}_i,
\end{equation}
where $n$ is the number of utterances in the dialogue.

\begin{equation}
{p}^{rel}_{\varepsilon_i,\varepsilon_j} = MLP_{rel}(r^\varepsilon_{i,j}).
\end{equation}

The quadruple $(t, a, o, p)$ is a successful quad prediction if and only if ${p}^{rel}_{t,a} = \textit{R}$, ${p}^{rel}_{t,o} = \textit{R}$, and $argmax({p}^{rel}_{a,o}) \in \{\textit{Pos}, \textit{Neg}, \textit{Neu}\}$.

\subsection{B.2 Sub-dialogue level Quad Matching}
In the last step, we extracted all candidate target, aspect, and opinion entity spans from the dialogue.
In this phase (Figure \ref{fig:1} B.2), we match them at the sub-dialogue level.
Based on the previously partitioned set of sub-dialogues $\mathcal{C}$, we group entities that belong to the same sub-dialogue together.
Therefore, for a sub-dialogue $c \in \mathcal{C}$, its candidate target, aspect, and opinion entity spans are denoted as: 
$\mathcal{E}^t=\{t_1, ..., t_{|\mathcal{E}^t|}\}$, $\mathcal{E}^a=\{a_1, ..., a_{|\mathcal{E}^a|}\}$ and $\mathcal{E}^o=\{o_1, ..., t_{|\mathcal{E}^o|}\}$, respectively.

\subsubsection{Quad Label Schema}
These three types of candidate entities can form three types of pairs: \{target-aspect (\textit{TA}), target-opinion (\textit{TO}), aspect-opinion (\textit{AO})\}.
Therefore, we iterate through each of these types and classify their relationships in \{Related (\textit{R}), Positive (\textit{Pos}), Negative (\textit{Neg}), Neutral (\textit{Neu}), Invalid (\textit{Inv})\}. 
For a candidate sentiment triplet (target, aspect, opinion) traversed from $\mathcal{E}^t$, $\mathcal{E}^a$ and $\mathcal{E}^o$, the quadruple (target, aspect, opinion, sentiment) can only be successfully confirmed when the \textit{TA} and \textit{TO} pairs are predicted as \textit{R} (which confirms that they belong to the same tuple), and the \textit{AO} pairs are predicted into one of \{\textit{Pos}, \textit{Neg}, \textit{Neu}\} (which confirms the sentiment polarity of the tuple).



\subsubsection{Quad Matching}
\label{quad_match}
For two spans $\varepsilon_i$ and $\varepsilon_j$ that belong to the same sub-dialogue $c$ ($\varepsilon_i, \varepsilon_j \in \{\mathcal{E}^t, \mathcal{E}^a, \mathcal{E}^o\}$) and not to the same entity category ($y^{ent}_i \neq y^{ent}_j$), We can obtain the matching relationship between them through the following process:
First, their span representations can be obtained by a max-pooling operation of their span range in $\bm{H}^{c}$:
\begin{equation}
h^\varepsilon_i = Pool(\bm{H}^{c}_{[\varepsilon_i]}), h^\varepsilon_j = Pool(\bm{H}^{c}_{[\varepsilon_j]}),
\label{eq166}
\end{equation}
where  $\bm{H}^{c} \subseteq \bm{H}$ is the representations of all utterances in $c$.
The ($\varepsilon_i$, $\varepsilon_j$) pair representation $r^\varepsilon_{i,j} \in \mathbb{R}^{3d+\eta}$ is formulated as:
\begin{equation}
r^\varepsilon_{i,j} = h^\varepsilon_i \oplus h^\varepsilon_j \oplus Pool(\bm{H}^{c}_{[\varepsilon_i:\varepsilon_j]}) \oplus Biaf(h^\varepsilon_i, h^\varepsilon_j),
\end{equation}
\begin{equation}
Biaf(h^\varepsilon_i, h^\varepsilon_j) = {h^\varepsilon_i}^T \mathbf{W_2} h^\varepsilon_j,
\label{eq16}
\end{equation}
where $\mathbf{W_2} \in \mathbb{R}^{d \times \eta \times d}$ is a learnable parameter matrix.
Then, we employ a label-wise linear layer $MLP_{rel}: \{\mathbb{R}^{3d+\eta} \rightarrow \mathbb{R}^{5} \}$ to predict the label ${p}^{rel}_{\varepsilon_i,\varepsilon_j} \in \{ \textit{R}, \textit{Pos}, \textit{Neg}, \textit{Neu}, \textit{Inv}\}$ of pair ($\varepsilon_i$, $\varepsilon_j$):
\begin{equation}
{p}^{rel}_{\varepsilon_i,\varepsilon_j} = MLP_{rel}(r^\varepsilon_{i,j}).
\label{eq19}
\end{equation}

The quadruple $(t, a, o, p)$ is a successful quad prediction if and only if ${p}^{rel}_{t,a} = \textit{R}$, ${p}^{rel}_{t,o} = \textit{R}$, and $p=argmax({p}^{rel}_{a,o}) \in \{\textit{Pos}, \textit{Neg}, \textit{Neu}\}$. 
%

The loss of the quadruple matching in sub-dialogue $c$ is formulated as:
\begin{equation}
l^{rel}_c = -\frac{1}{m} \sum_{y^{ent}_i \neq y^{ent}_j} \log ({p}^{rel}_{i,j} I({y}^{rel}_{i,j})).
\end{equation}
The sum of the losses $l^{rel}$ for all sub-dialogues is calculated as: $\mathcal{L}_{ent} = \sum_{i=1}^{|\mathcal{C}|} l^{rel}_i$.
The overall loss of our SEMDia is:
\begin{equation}
\mathcal{L} = \mathcal{L}_{ent} + \mathcal{L}_{rel}.
\end{equation}

%% file: 5-experiments.tex
\section{Experiments}\label{sec:experiments}








\subsection{Datasets}
Following previous works, we conduct our experiments on the DiaASQ datasets \cite{li-etal-2023-diaasq}, which contain English and Chinese versions collected from the smartphone domain.
The Chinese version includes a total of 1,000 dialogues, 7,452 utterances, and 5,742 sentiment quadruples, whereas the English version contains 5,514 sentiment quadruples. On average, each dialogue involves around five speakers. The dataset also includes 1,275 cross-utterance quadruples in Chinese (22.2\%) and 1,227 in English (22.3\%). 
The statistics are provided in Table \ref{tab:statistics}.

\subsection{Baselines}
Due to limited related research, we follow previous works \cite{li-etal-2023-diaasq, 10.1145/3589334.3645355}, which compared several high-performing sentence-level ABSA models. These models were adapted to complete the DiaASQ task, categorized into three types:
\begin{itemize}[]
    \item Tagging-based Methods: \textbf{CRFExtract} \cite{cai-etal-2021-aspect} is a sentence-level quadruple extraction model based on a CRF layer and BIO tagging strategy in a pipeline.
    \textbf{GTS} \cite{wu-etal-2020-grid} proposes a grid tagging scheme to fully capture the associations between token pairs for end-to-end extraction.
    \textbf{BDTF} \cite{xu-etal-2021-learning} further addresses the boundary dependency issue based on this scheme.

    \item Span-based Methods: 
    \textbf{SpERT} \cite{eberts2020span} introduces a span-based transformer for sentence-level joint extraction. \textbf{Span-ASTE} \cite{xu-etal-2021-learning} proposes a dual-channel span pruning strategy combining multi-task supervision.

    \item Generative Method: \textbf{ParaPhrase} \cite{zhang-etal-2021-aspect-sentiment} is a generative seq-to-seq model for sentence-level quadruple extraction.
\end{itemize}

In addition to the work above, we also compared models specifically designed for the DiaASQ task:
\begin{itemize}[]
    \item \textbf{Meta-WP} \cite{li-etal-2023-diaasq} is a dialogue-level quadruple extraction model with a end-to-end table tagging manner.
    \item \textbf{H2DT} \cite{Li2024HarnessingHD} is a dialogue-level quadruple extraction model based on dialogue graph and GCN enhancement.
    \item \textbf{DMCA} \cite{10447873} introduces a multi-scale aggregation network to fully aggregate contextual information.
    \item \textbf{DMIN} \cite{huang-etal-2024-dmin} emploies a multi-granularity integrator to enhance token-level contextual understanding.
    \item \textbf{STS} \cite{10.1145/3589334.3645355} proposes a cross-pair interaction and tagging method to address the complex interaction modeling in long dialogue sequences.
\end{itemize}

Recent large-scale language models (LLMs) have achieved impressive results in some NLP tasks. Although they still struggle with ABSA tasks \cite{zhang-etal-2024-sentiment}, we include them in our comparison.
\begin{itemize}[]
    \item LLMs: \textbf{GPT-4o}\footnote{https://chat.openai.com} and \textbf{DeepSeek-V3}\footnote{https://www.deepseek.com}. We evaluated their performance under the four-shot setting.
\end{itemize}
We designed a unified prompt template for these LLMs to guide them in completing the DiaASQ task. The prompt template includes both instructions and demonstrations (for few-shot settings). The instruction is as follows:

\textit{Given a dialogue: <Input>, please extract all (Target, Aspect, Opinion, Sentiment) quads. Don't explain yourself, your answer:}

Additionally, we incorporated annotations for each specific element in the quads within the template to enhance the LLMs' understanding of the task. 
To further leverage the in-context learning capability of the LLMs, we randomly selected four data samples from the training set as demonstrations in the four-shot setting.

\begin{table}[t]
\centering
\belowrulesep=0pt
\aboverulesep=0pt
\tabcolsep=8pt
\caption{Statistics of the DiaASQ datasets. where Dia., Utt. and Spk. donate the number of dialogues, utterances, and speakers, respectively. Tgt., Asp., and Opi. donate the number of target/aspect/opinion terms, respectively. t-a., t-o., and a-o. donate the number of corresponding pairs, respectively. Intra., Cross. donates the number of intra-/cross-utterance quadruples.}
\scalebox{1}{
\begin{tabular}{ccccccc}
\Xhline{0.9pt}
       & \multicolumn{3}{c}{ZH} & \multicolumn{3}{c}{EN} \\
       \cmidrule(lr){2-4}\cmidrule(lr){5-7}
       & train  & valid  & test & train  & valid  & test \\ \hline
\multicolumn{1}{c|}{Dia.}   & 800    & 100    & 100  & 800    & 100    & 100  \\
\multicolumn{1}{c|}{Utt.}   & 5,947  & 748    & 757  & 5,947  & 748    & 757  \\
\multicolumn{1}{c|}{Spk.}   & 3,986  & 502    & 503  & 3,986  & 502    & 503  \\ \hline
\multicolumn{1}{c|}{Tgt.}   & 6,652  & 823    & 833  & 6,613  & 822    & 829  \\
\multicolumn{1}{c|}{Asp.}   & 5,220  & 662    & 690  & 5,109  & 644    & 681  \\
\multicolumn{1}{c|}{Opi.}   & 5,622  & 724    & 705  & 5,523  & 719    & 691  \\ \hline
\multicolumn{1}{c|}{t-a.}    & 4,823  & 621    & 597  & 4,699  & 603    & 592  \\
\multicolumn{1}{c|}{t-o.}    & 6,062  & 758    & 767  & 5,931  & 750    & 751  \\
\multicolumn{1}{c|}{a-o.}    & 4,297  & 538    & 523  & 3,989  & 509    & 496  \\ \hline
\multicolumn{1}{c|}{Quad.}  & 4,607  & 577    & 558  & 4,414  & 555    & 545  \\
\multicolumn{1}{c|}{Intra.} & 3,594  & 440    & 433  & 3,442  & 423    & 422  \\
\multicolumn{1}{c|}{Cross.} & 1,013  & 137    & 125  & 972    & 132    & 123  \\ \Xhline{0.9pt}
\end{tabular}
}
\label{tab:statistics}
\end{table}

\begin{table*}[th]
\centering
\belowrulesep=0pt
\aboverulesep=0pt
\tabcolsep=6pt
\caption{Main results on the DiaASQ datasets. 
$\dagger$ and $\ddagger$ denote the results are retrieved from  \citet{Li2024HarnessingHD} and \cite{10.1145/3589334.3645355}. Other results are retrieved from the original paper. $\sharp$ denotes the results of our rerun.
The best results are highlighted in \textbf{bold} and the second best are \underline{underlined}. }
\begin{tabular}{ccccccccccccc}
\Xhline{0.9pt}
\multirow{2}{*}{Data} & \multirow{2}{*}{Methods} & \multicolumn{3}{c}{Entity(F1)} & \multicolumn{3}{c}{Pairs(F1)} & \multicolumn{3}{c}{Triplet} & \multicolumn{2}{c}{Quadruple} \\
\cmidrule(lr){3-5}\cmidrule(lr){6-8}\cmidrule(lr){9-11}\cmidrule(lr){12-13}
&                          & \textit{T}        & \textit{A}        & \textit{O}        & \textit{T-A}      & \textit{T-O}      & \textit{A-O}     & P       & R       & F1         & 	Iden-F1        & Micro-F1       \\ \hline 
\multirow{16}{*}{EN}   
& CRF-Extract$^\dagger$              & 88.31    & 71.71    & 47.90    & 34.31    & 20.94    & 19.21   & /       & /       & 12.80       & /        & 11.59   \\ 
& BDTF$^\ddagger$              & 88.60 &73.37 &62.53   & 49.26 &47.55 &49.95   & /       & /       &     & 38.80 &31.81  \\
& GTS$^\ddagger$              & 88.62    & 74.71    & 60.22    & 47.91    & 45.58    & 44.27   & /       & /       & /       & 36.80       & 33.31   \\ 
& SpERT$^\dagger$                    & 87.82    & 74.65    & 54.17    & 28.33    & 21.39    & 23.64   & /       & /       & 13.38      & /        & 13.07   \\
& Span-ASTE$^\dagger$                & /        & /        & /        & 42.19    & 30.44    & 45.90   & /       & /       & 28.34         & /        & 26.99   \\
& ParaPhrase$^\dagger$               & /        & /        & /        & 37.22    & 32.19    & 30.78   & /       & / & 26.76       & /        & 24.54   \\
\cmidrule(lr){2-13}
& GPT-4o$_{\text{four-shot}}$    & 79.91   & 64.24  & 57.28   & 34.64  & 37.76  & 27.50   & 24.90  & 21.34  & 22.98   & 14.78  & 16.94   \\
& DeepSeek-V3$_{\text{four-shot}}$    & 80.54   & 65.10  & 58.77   & 35.99  & 39.05  & 29.66   & 25.15  & 22.66  & 23.84   & 15.15  & 17.38   \\
\cmidrule(lr){2-13}
& Meta-WP    & 88.62    & 74.71    & 60.22    & 47.91    & 45.58    & 44.27   & /       & /       & 36.80         & /        & 33.31   \\
& H2DT-large   & 88.69    & 73.81    & 62.61    & 48.69    & 48.84  & 52.47   & \underline{44.36}   & \underline{40.23}   & \underline{42.19}   & 37.20    & 39.01   \\
& DMCA   & 88.11   &73.95   & 63.47  & 53.08    & 50.99  & 52.40  & /   & /   & /  & /   & 37.96  \\
& DMIN     & /     &/   & /   &53.49   & 52.66   & 52.09  & /   & /   & /  & /   & 39.22  \\
& STS$^\sharp$      & \underline{89.00}   &\underline{74.55}  & \underline{63.51} & \underline{54.20}    & \underline{52.82}    & \underline{54.64}  & /   & /   & /  & \underline{39.09}  & \underline{40.45}  \\
& SEMDia (Ours)   & \textbf{89.12} & \textbf{75.01}   & \textbf{64.07} & \textbf{55.45}  & \textbf{53.22}   & \textbf{56.59}    &  \textbf{50.63} & \textbf{43.27}  & \textbf{46.66} & \textbf{39.39}  & \textbf{42.40} \\ 
\hline
                      \rowcolor[HTML]{EFEFEF}
& $\Delta$  &  \textbf{0.12}$\uparrow$    & \textbf{0.46}$\uparrow$   & \textbf{0.56}$\uparrow$   & \textbf{1.25}$\uparrow$   & \textbf{0.40}$\uparrow$   & \textbf{1.95}$\uparrow$   & \textbf{6.27}$\uparrow$  &   \textbf{3.04}$\uparrow$ &  \textbf{4.47}$\uparrow$   & \textbf{0.30}$\uparrow$   & \textbf{1.95}$\uparrow$     \\ 
                      \hline
\multirow{16}{*}{ZH}   
& CRF-Extract$^\dagger$  & 91.11    & 75.24    & 50.06    & 32.47    & 26.78    & 18.90   & /       & /       & 9.25     & /        & 8.81    \\
& BDTF$^\ddagger$        & 91.08 &76.24 &60.88    & 51.41 &49.33 &52.58 & /       & /       & 9.25     & 41.06 &34.22  \\
& GTS$^\ddagger$              & 90.23 &76.94 &59.35    &48.61 &43.31 &45.44   & /       & /       & /       & 37.51 &34.94   \\ 
& SpERT$^\dagger$                    & 90.69    & 76.81    & 54.06    & 38.05    & 31.28    & 21.89   & /       & /       & 14.19    & /        & 13.00   \\
& Span-ASTE$^\dagger$                & /  & /        & /        & 44.13    & 34.46    & 32.21   & /       & /       & 30.85   &/      & 27.42   \\
& ParaPhrase$^\dagger$               & /        & /        & /        & 37.81    & 34.32    & 27.76   & /       & /       & 27.98   & /        & 23.27   \\
\cmidrule(lr){2-13}
& GPT-4o$_{\text{four-shot}}$    & 82.56   & 67.22  & 60.04   & 37.80  & 40.08  & 30.14   & 27.73  & 24.47  & 26.00   & 18.60  & 20.67   \\
& DeepSeek-V3$_{\text{four-shot}}$    & 84.46   & 69.21  & 62.74   & 40.25  & 43.69  & 34.59  & 29.43  & 26.78  & 28.95   & 19.26  & 21.16  \\
\cmidrule(lr){2-13}
& Meta-WP                  & 90.23    & 76.94    & 59.35    & 48.61    & 43.31    & 45.44   & /       &/       & 37.51   &/           & 34.94   \\
& H2DT     & 91.72    & 76.93    & \underline{61.87}    & 50.48    & 48.80    & 52.40   & \underline{45.40}   & \underline{40.50}   & \underline{42.81}   & 38.17    & 40.34  \\
& DMCA       & \textbf{92.03}   & \underline{77.07}   & 60.27   & \underline{56.88}   & \underline{51.70}  & 52.80  & /   & /   & /  & /   & 42.68  \\
& DMIN                     & /     &/   & /   & \textbf{57.62}    & 51.65   & \underline{56.16}  & /   & /   & /  & /    & \underline{44.49}  \\
& STS$^\sharp$      &  91.32   &76.87  & 60.87 & 53.49    & 50.01    & 53.34  & /   & /   & /  & \underline{42.26}  & 40.08  \\
& SEMDia (Ours)   & \underline{91.90} &   \textbf{77.25}  & \textbf{64.70}      &  56.14  & \textbf{51.78}   &   \textbf{56.22} &\textbf{51.23}  & \textbf{42.35} & \textbf{46.37} &    \textbf{42.44} &  \textbf{44.61}  \\ \hline
                      \rowcolor[HTML]{EFEFEF}
& $\Delta$   &  0.13$\downarrow$  &  \textbf{0.18}$\uparrow$   &  \textbf{2.83}$\uparrow$  & 1.40$\downarrow$  & \textbf{0.08}$\uparrow$  & \textbf{0.08}$\uparrow$ & \textbf{5.83}$\uparrow$ &  \textbf{1.85}$\uparrow$ & \textbf{3.99}$\uparrow$ &\textbf{0.18}$\uparrow$&\textbf{0.12}$\uparrow$  \\ \Xhline{0.9pt}
\end{tabular}
\label{tab:mainresult}
\end{table*}

\subsection{Experimental Settings}
\subsubsection{Implementation}
For the English dataset, we initialize the PLM with Roberta-base \cite{liu2019roberta} in contrast to the previous SOTA model \citet{Li2024HarnessingHD}, which used the Roberta-large version.
For the Chinese dataset, we initialize the PLM with Chinese-Roberta-wwm-ext \cite{cui2021pre}.
We conduct pipeline training for our model. 
In the Span Extraction and Quad Matching modules, we use separate PLMs, which will yield marginal benefits for multitask learning.
The hidden layer dimension $d$ of the PLM is $768$. 
For the MLP dimension $d'$ in Formula \ref{eq7}, it is set to $256$. 
$\eta$ in parameter matrices $\mathbf{W_1}$ and $\mathbf{W_2}$ are set to $64$. 
The parallel merge ratio $\sigma$ in the DSEM algorithm is set to $0.15$ for the EN dataset and $0.35$ for the ZH dataset.
We employ the AdamW optimizer with an initial learning rate of 1e-5 and weight decay of 1e-8.
We first train the Span Extraction module for 10 epochs and then train the Quad Matching module for another 5 epochs.
The model is trained on RTX 8000 GPUs with a batch size of 2.
In each epoch, we evaluate the model on the validation set and save the best one.
The results are averaged over 5 runs with random seeds.

\subsubsection{Metrics}
For entity, pair, and triplet extraction tasks, we used exact-F1 as the evaluation metric. 
For quadruple extraction, we used identification-F1 (iden-F1) \cite{barnes-etal-2021-structured} and micro-F1 as the evaluation metrics.
Iden-F1 does not consider sentiment polarity during computation, making it more suitable for evaluating the model's element-matching capability. While Micro-F1 measures the model's overall performance.


\subsection{Main Results}\label{rq1}
The primary results of our experiments are shown in Table \ref{tab:mainresult}.
We have the following observations:
Our SEMDia almost achieves the best performance across four tasks in both the EN and ZH datasets, showcasing its superiority in the DiaASQ task and its subtasks. 
In the English DiaASQ (\textbf{Quadruple Extraction}) task, our SEMDia significantly outperforms the leading baseline STS $0.3\%$ in Iden-F1 and $1.95\%$ in Micro-F1, while STS uses the large version of Roberta, but we only use the base version.
In the Chinese dataset, our SEMDia outperforms STS $0.18\%$ in Iden-F1 and $4.53\%$ in Micro-F1.

In the \textbf{Triplet Extraction} task, our model's improvements are primarily reflected in precision.
Specifically, the gains in precision are $6.27\%$ and $5.83\%$ on the EN and ZH datasets, respectively.
While the gains in recall are $3.04\%$ and $1.85\%$, respectively.
These improvements can be attributed to the partitioning of sub-dialogue, which reduces the matching space for sentiment elements, thereby significantly increasing precision.
In the \textbf{Entity Extraction} task, it can be observed that our model shows greater improvement on the more challenging \textbf{opinion extraction} tasks, which demonstrates the effectiveness of our Utterance-level Span Extraction module.
Overall, due to the bucket effect, these improvements promise more significant gains for downstream composite tasks.
We also explore the performance of LLMs on the DiaASQ task, where they struggle to achieve competitive results, which is consistent with the observations of \citet{huang-etal-2024-dmin} and \citet{10.1145/3589334.3645355}.

\begin{table*}[t]
\centering
\tabcolsep=2.5pt
\belowrulesep=0pt
\aboverulesep=0pt
\caption{Ablation results of three matching tasks.}
\begin{tabular}{lcccccccccccccccc}
\Xhline{0.9pt}
\multirow{3}{*}{Methods} & \multicolumn{8}{c}{EN}                                                            & \multicolumn{8}{c}{ZH}                                                            \\ \cmidrule(lr){2-9} \cmidrule(lr){10-17} 
                         & \multicolumn{3}{c}{Pairs} & \multicolumn{3}{c}{Triplet} & \multicolumn{2}{c}{Quadruple} & \multicolumn{3}{c}{Pairs} & \multicolumn{3}{c}{Triplet} & \multicolumn{2}{c}{Quadruple} \\ \cmidrule(lr){2-4}\cmidrule(lr){5-7} \cmidrule(lr){8-9}  \cmidrule(lr){10-12}\cmidrule(lr){13-15}\cmidrule(lr){16-17} 
& T-A     & T-O    & A-O    & P      & R      & F1      & Iden-F1     & Micro-F1     & T-A     & T-O    & A-O    & P      & R      & F1      & Iden-F1     & Micro-F1    \\ \hline
SEMDia                   & \textbf{55.45} &\textbf{53.22} &\textbf{56.59}   &\textbf{50.63} &\textbf{43.27}&\textbf{46.66} &\textbf{39.39} &\textbf{42.40} &  \textbf{56.14} & \textbf{51.78} & \textbf{56.24} & \textbf{51.23} & \textbf{42.35} & \textbf{46.37} & \textbf{42.44} & \textbf{44.61}  \\ \cline{1-1}
- w/o \textit{sub-dia}            &54.90 &52.37 &55.11   &  49.61  & 42.21 & 45.61   &  38.52  &  41.78  &  55.79  &  51.64   & 55.72 & 50.57   &  41.50  & 45.59   &  41.75   &  44.05   \\
- w/ \textit{Reply}               & 54.66 &52.21 &55.38 & 49.39   &  42.16  & 45.49  &  38.28    &  41.49   & 55.32 &51.52 &55.92   & 50.24   & 41.44  &  45.42  &  41.35  &  43.81      \\
- w/ \textit{K-means}             &   54.22 &51.12 &53.70  &  48.56  &  41.02  &  44.47  &     37.67 & 40.58   &  54.84 &53.52 &53.27   &   49.31     & 41.54  &  45.09   &    40.89   &  43.05     \\ \Xhline{0.9pt}
\end{tabular}
\label{tab:ablation}
\end{table*}

\subsection{Ablation Study}
\label{ablation}
To validate the proposed method, we conduct our ablation experiments with the following settings:

\textbf{1) -w/o \textit{sub-dia}}: We remove the partitioning strategy and directly extract over the entire dialogue.

\textbf{2) -w/ \textit{Reply}}: We remove the DSEM algorithm, replacing it with direct partitioning using reply relations (i.e., grouping all utterances in the same reply line into one cluster).

\textbf{3) -w/ \textit{K-means}}: We replaced the DSEM algorithm with K-means clustering. Note that K-means requires pre-setting the number of clusters, which we set to the optimal value of 3 here.

We do not present the entity extraction results, as it is at the utterance level and unaffected by ablation.
As shown in Table \ref{tab:ablation}, when removing the sub-dialogue partitioning strategy (-w/o \textit{sub-dia}), the model's performance decreases across all tasks.
This demonstrates that using a sub-dialogue partitioning strategy is crucial.
When replacing the DSEM algorithm with the reply relation (-w/ \textit{Reply}), the model's performance shows a more significant decline. 
This is mainly because partitioning based on reply relations causes many cross-utterance quadruples to be distributed across different sub-dialogues, resulting in decreased performance.
When using K-means, the model's performance suffers a significant decline because the predefined number of clusters in K-means can lead to more erroneous partitions.

\begin{table}[]
\centering
\tabcolsep=2pt
\caption{Time complexity and computing efficiency. The $N$ and $K$ denote the number of tokens and enumerated spans in span-based methods, where $K\lesssim N$.
The $L$ and $C$ denote the number of utterances and extracted entities in our method, where $L \ll N$ and $C \ll K$.
Avg-F1 presents the average Micro-F1 on the two datasets, and Infer. denotes the inference time per sample.}
\begin{tabular}{lccc}
\hline
Model         & Time Complexity & Avg-F1 & Infer. ($\backslash$ ms) \\ \hline
GTS           &   $O(N^2)$    &  34.13   &    6.03    \\
Span-ASTE     &  $O(N^2)+O(K^3)$    &   27.21  &  6.75        \\
STS           &  $O(N^2)+O(K^2)$  &  40.27   &    6.22     \\ \hline
SEMDia (Ours) &  $O(L^3)+O({(\frac{N}{L})}^2)+O(C^3)$   &  \textbf{43.51}   &   \textbf{4.91}   \\ \hline
\end{tabular}
\label{computing}
\end{table}

\subsection{Computing Efficiency}
\label{rq-computing}
To gain a deeper understanding of SEMDia's performance, we further examined the time complexity of our approach in comparison to baseline methods, as well as the computational efficiency. 
The results are presented in Table \ref{computing}.
Since $K \lesssim N$, so for Span-ASTE, it's time complexities $O(N^2)+O(K^3) \lesssim O(N^3)$; for STS, it's time complexities $O(N^2)+O(K^2) \lesssim O(N^2)$.
For our SEMDia, as shown in Algorithm \ref{xxx}, the time complexity of the dialogue partitioning phase is $O(L^3)$. Afterward, due to the partitioning into sub-dialogues, the time complexity of the entity extraction phase is $O({(\frac{N}{L})}^2)$. 
Finally, as described in section \ref{quad_match}, the time complexity of the quadruple matching phase is $O(C^3)$.
Since $L \ll N$ and $C \ll N$, so the time complexities of our SEMDia is $O(L^3)+O({(\frac{N}{L})}^2)+O(C^3) \lesssim O(N^2/L^2)$, which is significantly lower than other methods.
The presented Avg-F1 score and inference time indicate that SEMDia achieves the best performance with lower computational cost, demonstrating the efficiency and effectiveness of our method.

\begin{figure}[t]
    \centering
    \includegraphics[width=0.45\textwidth]{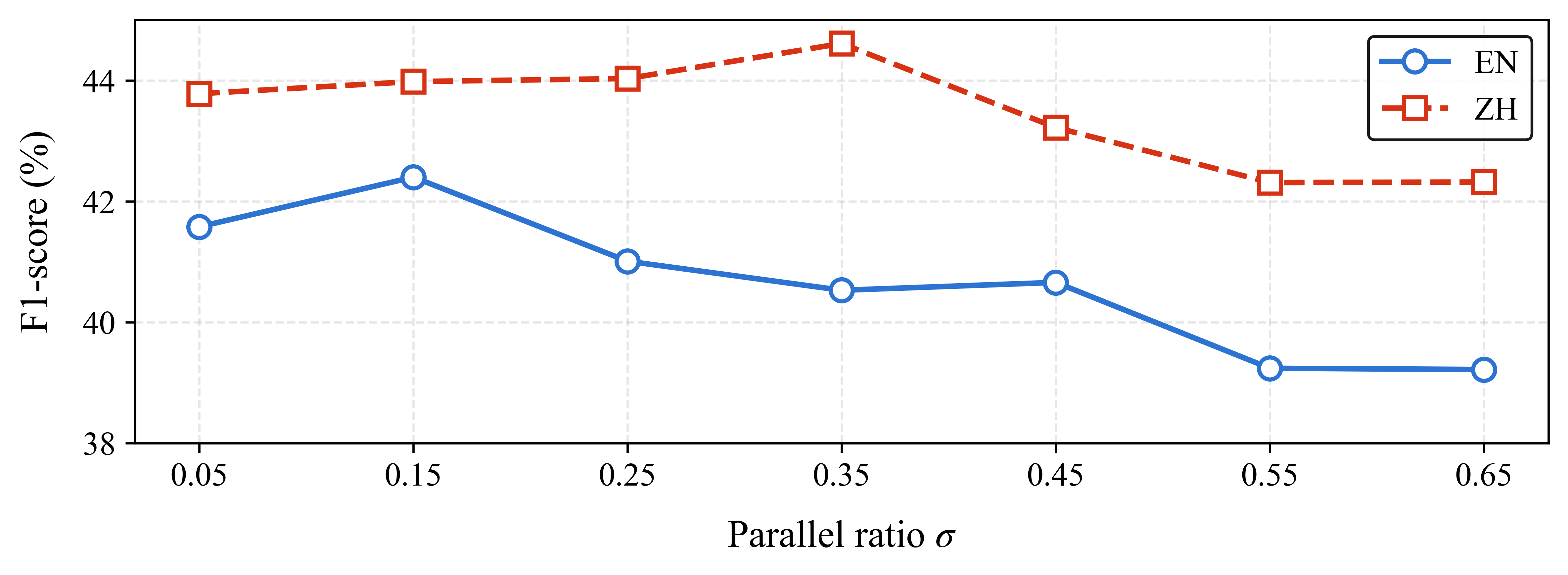}
    \vspace{-0.3cm}
    \caption{Sensitivity of parallel merge ratio $\sigma$.}
    \label{fig:2}
\end{figure}

\begin{figure*}[t]
    \centering
    \includegraphics[width=0.9\textwidth]{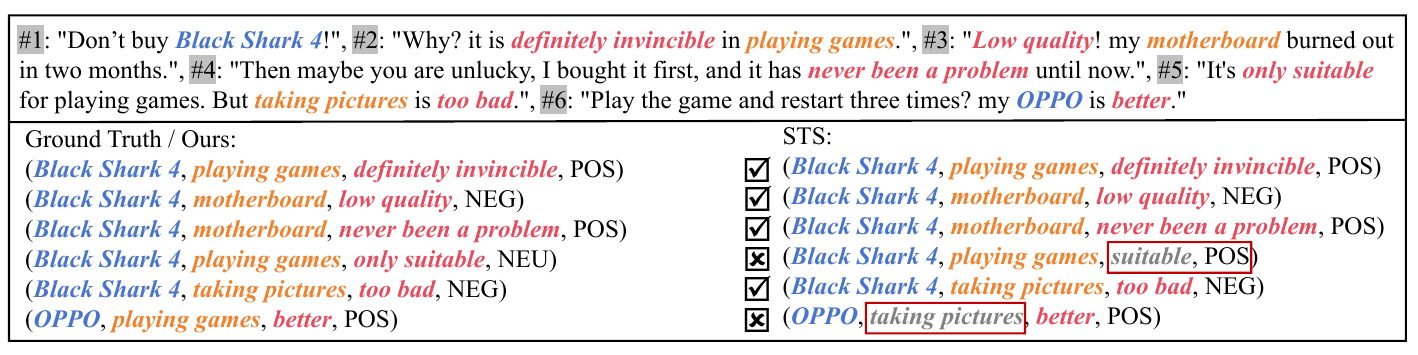}
    \vspace{-0.3cm}
    \caption{Case study. }
    \label{fig:5}
\end{figure*}

\subsection{Effect of Parallel Merge Ratio $\sigma$}
\label{rq2}
The hyperparameter $\sigma$ in the DSEM algorithm controls the number of clusters that can be merged in each iteration round.
To investigate the impact of $\sigma$, we conduct a sensitivity experiment.
As shown in Figure \ref{fig:2}, we plot the Micro-F1 scores for quadruple extraction at different $\sigma$ values. 
Due to the varying optimal 2D encoding trees corresponding to changes in $\sigma$, there are differences in partitioning and F1 score. 
It can be observed that the model performs best on the English dataset when $\sigma$ is 0.15, and on the Chinese dataset when $\sigma$ is 0.35.
This occurs because choosing a larger $\sigma$ can result in the merging of suboptimal clusters. 
Conversely, a smaller $\sigma$ reduces the efficiency of merging. 
When $\sigma$ exceeds 0.55, the parallel merging operation count reaches its upper limit, and the optimal 2D encoding tree no longer varies with $\sigma$.

\subsection{In-depth Study on Complex Scenarios}
\label{rq3}
\subsubsection{Performance on Cross-utterance Quadruples}
\begin{figure}[t]
    \centering
    \hspace{-0.4cm}
    \includegraphics[width=0.45\textwidth]{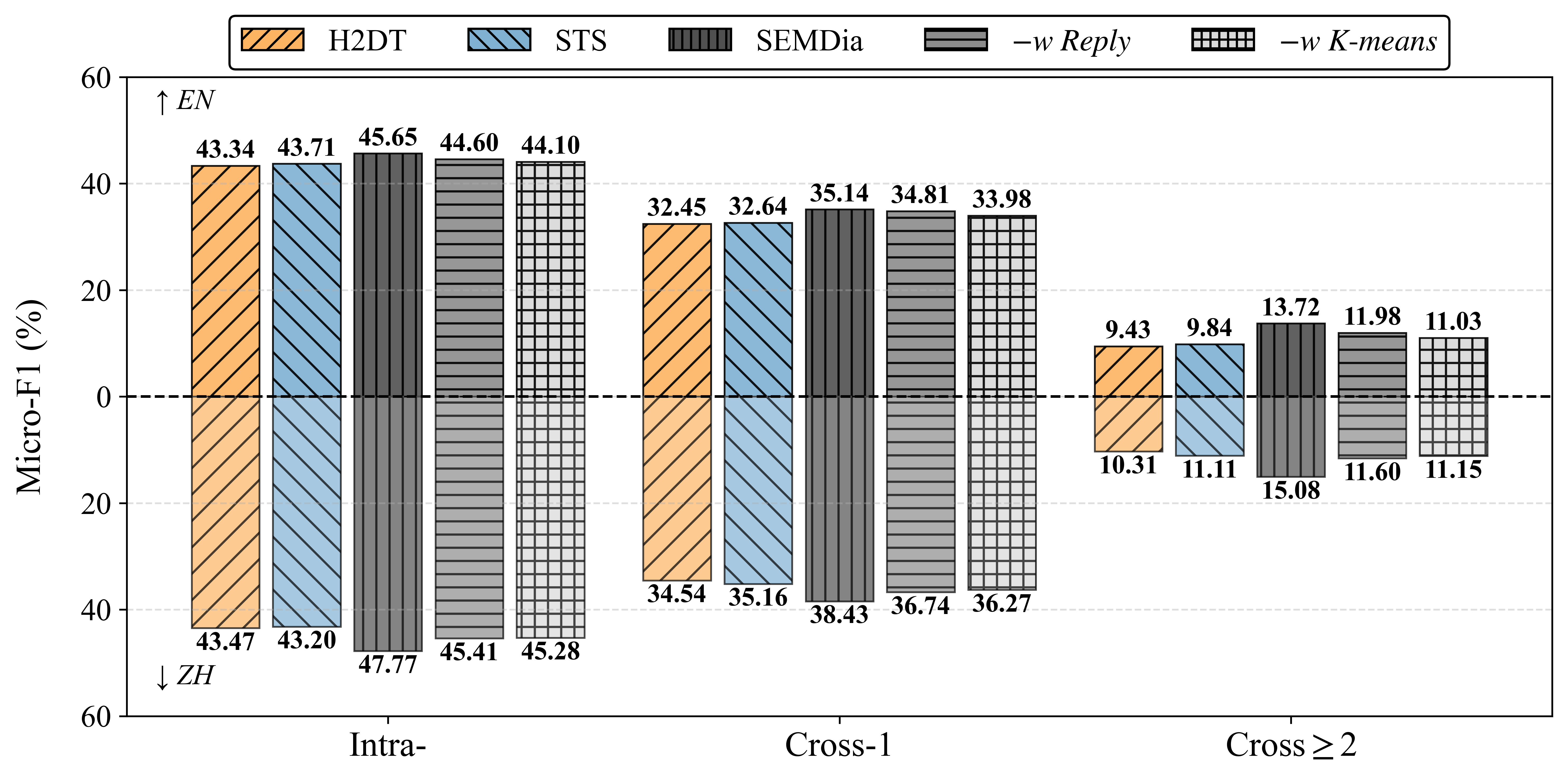}
    \vspace{-0.2cm}
    \caption{Performance on intra-, cross-1-, and cross$\ge$2-utterance quadruple extraction of different methods.}
    \label{fig:3}
\end{figure}

We conduct an in-depth analysis of the performance of different methods at varying quadruple distances and present the results in Figure \ref{fig:3}, where the distances are categorized as intra-utterance, cross-1-utterance, and greater than 2-utterances (cross$\ge$2-utterance).
Our SEMDia demonstrates significant improvements over baseline methods in both intra- and cross-utterance extraction tasks. 
Compared to STS, our model improves by 1.94\%, 2.5\%, and 3.88\% in intra-, cross-1-, and cross$\ge$2-utterance extraction on the EN dataset, and similar results are also demonstrated on the ZH dataset.
The more pronounced improvement in cross-utterance extraction demonstrates the effectiveness of our core idea of sub-dialogue partitioning.
Furthermore, when the sub-dialogue strategy is removed (-w/ \textit{Reply}), we observe a decrease in both intra- and inter-utterance. 
Nevertheless, the performance still surpasses that of H2DT, thereby affirming the effectiveness of our two-stage extraction-matching model.

\begin{figure}[t]
    \centering
    \includegraphics[width=0.45\textwidth]{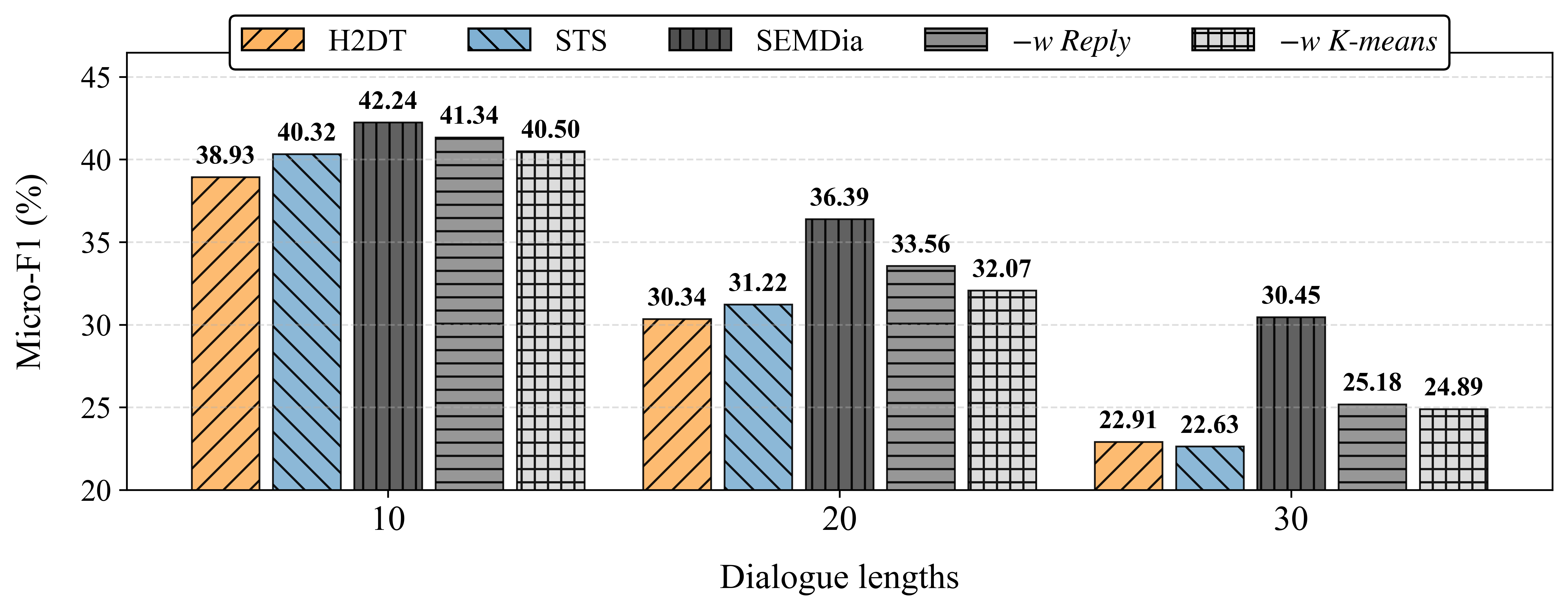}
    \vspace{-0.2cm}
    \caption{Performance on longer dialogues of varying lengths.}
    \label{fig:4}
\end{figure}

\subsubsection{Performance on Longer Dialogues}
\label{rq4}
As the length of the dialogue increases, the DiaASQ task becomes more challenging, which is exactly where our SEMDia excels. To thoroughly evaluate the model's performance in longer dialogues, we randomly selected 10 dialogues of length 10 from the original EN test dataset, and using the continuation capability of GPT-4o, we extended their lengths to 20 and 30, respectively. 
Afterward, we manually review and supplement the dialogues with quadruples. The final number of quadruples for each length is 64, 84, and 116, respectively.
We present the results in Figure \ref{fig:4}. As shown, with the increase in dialogue length, the performance of all methods declined to varying degrees. 
The longer the dialogue, the more significant the improvement of our SEMDia compared to the baselines, effectively demonstrating the superior handling capability of our method for long dialogues.


\subsection{Case Study}
\label{rq5}
To better highlight the advantages of our model, we provide a case study in Figure \ref{fig:5}. 
As shown, STS extracts the incomplete opinion \textit{suitable}, leading to incorrect predicted sentiment, while SEMDia effectively identifies the boundaries of the opinion words.
Additionally, STS incorrectly extracts (\textit{OPPO, taking pictures}) from the adjacent utterances \#5 and \#6, while our SEMDia avoids this error by partitioning them into separate sub-dialogues.

%% file: 6-conclusions.tex
\section{Conclusion}\label{sec:conclusion}
In this paper, we propose SEMDia for the DiaASQ task. 
By modeling dialogues into graphs, we efficiently partition them into semantically complete minimal sub-dialogues using a 2D structural entropy minimization algorithm. 
This unsupervised approach eliminates the noise unavoidably introduced when extracting quadruples at the entire dialogue scale.
Compared to error partition rates, this refined partitioning method noticeably enhances results. Furthermore, SEMDia utilizes a two-step extraction approach: initially extracting entities at the sentence level, followed by quadruple matching at the sub-dialogue level. Extensive experiments demonstrate SEMDia's significant utility in DiaASQ.
Exploitation of other dialogue-level information extraction tasks is left to future work.

\section*{Acknowledgments}
This research is supported by the National Key R\&D Program of China (No. 2023YFC3303800), NSFC through grants 62322202, 62441612, and 62476163, Beijing Natural Science Foundation through grant L253021, Local Science and Technology Development Fund of Hebei Province Guided by the Central Government of China through grant 246Z0102G, the “Pionee” and “Leading Goose” R\&D Program of Zhejiang through grant 2025C02044, Hebei Natural Science Foundation through grant F2024210008, and the Guangdong Basic and Applied Basic Research Foundation through grant 2023B1515120020.

\newpage
\section{GenAI Usage Disclosure}
In this manuscript, we have only used ChatGPT for writing refinement and grammar checking.
